%% file: main.tex
\documentclass[conference]{IEEEtran}
\IEEEoverridecommandlockouts
\usepackage{cite}
\usepackage{amsmath,amssymb,amsfonts}
\usepackage{algorithmic}
\usepackage{graphicx}
\usepackage{textcomp}
\usepackage{xcolor}
\usepackage{ulem}
\usepackage{tabularx, multirow, booktabs}
\def\BibTeX{{\rm B\kern-.05em{\sc i\kern-.025em b}\kern-.08em
    T\kern-.1667em\lower.7ex\hbox{E}\kern-.125emX}}

\usepackage[hidelinks]{hyperref}
\usepackage{multirow}
\usepackage{graphicx}
\usepackage{bbold}
\useunder{\uline}{\ul}{}
\usepackage{booktabs}
\usepackage{multirow}

\usepackage{hyperref} 
\usepackage{setspace}

\title{ALERTA-Net: A Temporal Distance-Aware Recurrent Networks for Stock Movement and Volatility Prediction}

\author{\textbf{Shengkun Wang}$^1$, \textbf{YangXiao Bai}$^2$, \textbf{Kaiqun Fu}$^2$, \textbf{Linhan Wang}$^1$,\textbf{Chang-Tien Lu}$^1$, \textbf{Taoran Ji}$^3$ \\
$^1${Department of Computer Science, Virginia Tech} \\
$^2${Department of Computer Science, South Dakota State University} \\
$^3${Department of Computer Science, Texas A\&M University - Corpus Christi} \\

\texttt{shengkun@vt.edu}, \texttt{bai.yangxiao@sdstate.edu}, 
 \texttt{kaiqun.fu@sdstate.edu}, \\  \texttt{linhan@vt.edu}, \texttt{ctlu@vt.edu},  \texttt{taoran.ji@tamucc.edu}
}
\makeatletter \def\@IEEEpubidpullup{8\baselineskip} \makeatother 

\usepackage{fancyhdr}
\usepackage{kantlipsum}
\fancypagestyle{plain}{
\fancyhf{}
\fancyhead[C]{Conference on \LaTeX} 
} \usepackage{eso-pic}

\begin{document}
\AddToShipoutPictureBG*{
    \AtPageUpperLeft{
        \setlength\unitlength{1in}
        \hspace*{\dimexpr0.5\paperwidth\relax}
        \makebox(0,-0.75)[c]{\textbf{2023 IEEE/ACM International Conference on Advances in Social Networks Analysis and Mining (ASONAM)}}
    }
}

\IEEEoverridecommandlockouts
\IEEEpubid{
\parbox{\columnwidth}{\vspace{-4\baselineskip} 
Permission to make digital or hard copies of part or all of this work for personal or classroom use is granted without fee provided that copies are not made or distributed for profit or commercial advantage and that copies bear this notice and the full citation on the first page. Copyrights for third-party components of this work must be honored. For all other uses, contact the Owner/Author.
\hfill\vspace{-0.8\baselineskip}\\ \begin{spacing}{1.2}
\small\textit{ASONAM '23}, November 6-9, 2023, Kusadasi, Turkey \\
\copyright\space Copyright is held by the owner/author(s). \\
ACM ISBN 979-8-4007-0409-3/23/11. \\ \url{https://doi.org/10.1145/3625007.3627488}
\end{spacing}
\hfill}
\hspace{0.9\columnsep}\makebox[\columnwidth]{\hfill}}
\IEEEpubidadjcol 

\maketitle
\begin{abstract}

For both investors and policymakers, forecasting the stock market is essential as it serves as an indicator of economic well-being. To this end, we harness the power of social media data, a rich source of public sentiment, to enhance the accuracy of stock market predictions. Diverging from conventional methods, we pioneer an approach that integrates sentiment analysis, macroeconomic indicators, search engine data, and historical prices within a multi-attention deep learning model, masterfully decoding the complex patterns inherent in the data. We showcase the state-of-the-art performance of our proposed model using a dataset, specifically curated by us, for predicting stock market movements and volatility.

\end{abstract}


\input{sections/introduction}

\input{sections/related_work}

\begin{figure}
    \centering
    \includegraphics[width=\columnwidth]{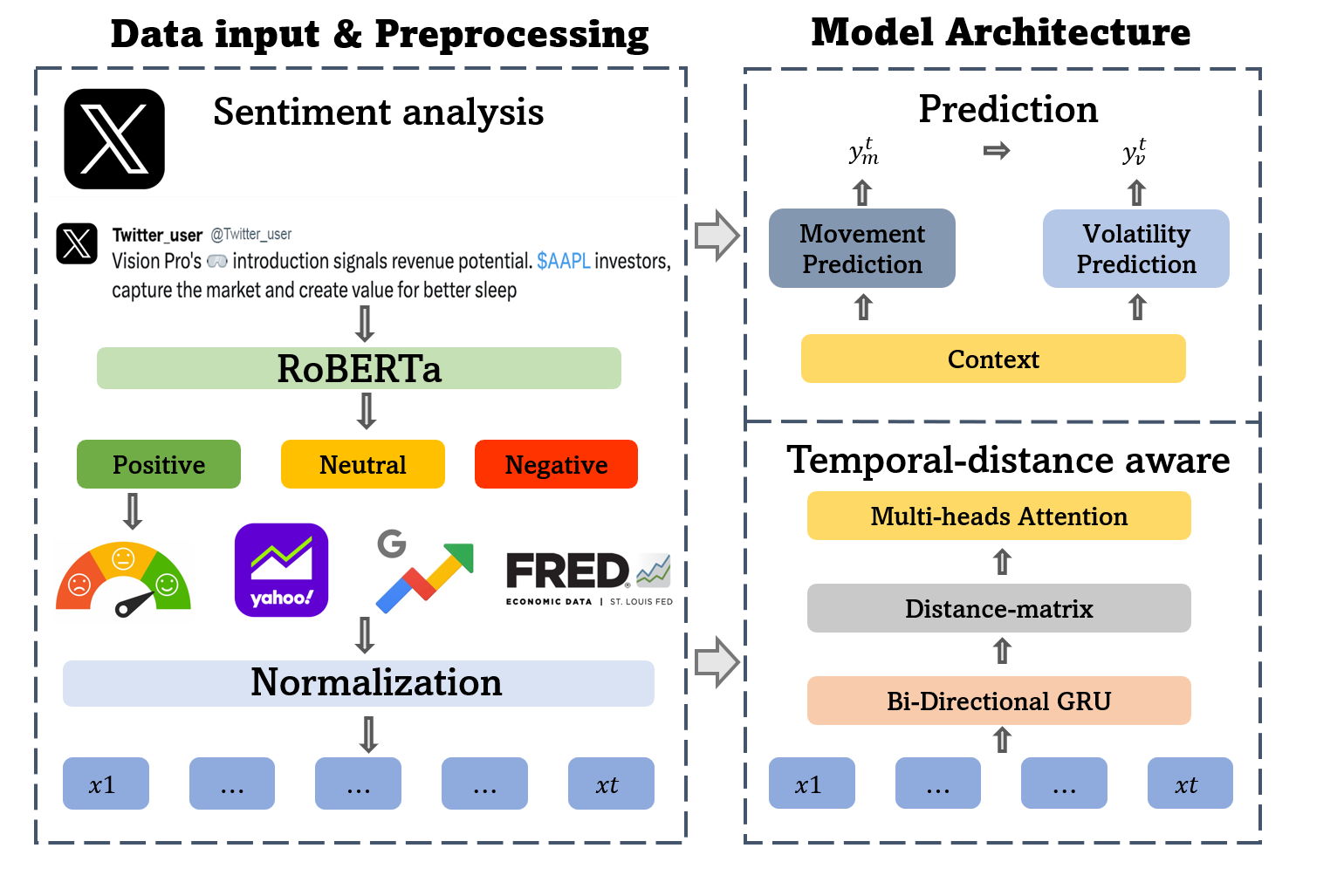}
    
    \label{fig:your_label}
    \par\vspace{5pt} 
    \parbox{\columnwidth}{
        Figure 1: The architecture of ALERTA-Net is designed to predict the movement \(y_m^{t}\), and volatility \(y_v^{t}\) on day \(t\). In the data input and preprocessing phase, we extract textual information from Twitter and convert it into sentiment scores; Then, ALERTA-Net utilizes these scores, along with other features, to make predictions, taking temporal distance into account.
    }
\end{figure}

\input{sections/prob_setup}

\input{sections/model}

\input{sections/experiment}

\input{sections/conclusion}

\bibliographystyle{IEEEtran}
\bibliography{ref.bib}

\end{document}

%% file: sections/introduction.tex
\section{Introduction}

Significantly influencing other business sectors~\cite{billah2016stock}, the stock market serves as a vital mechanism and is crucial for companies to raise capital. With U.S. stock holdings expected to hit \$40 trillion in 2023, equating to 1.5 times the nation's GDP, it stands as a major portion of the entire economy, highlighting the stock market's pivotal position as a benchmark for the U.S. economic landscape.
Our research centers on blue-chip stocks\footnoterule\footnote{Blue chip stocks are shares issued by financially robust, well-established companies with stellar reputations.}, which mirror the broader dynamics of the stock market.

We've selected 41 blue-chip stocks from 10 Global Industry Classification Standard (GICS)\footnoterule\footnote{GICS classifies companies into specific economic sectors and industry groups that most accurately represent their business operations.} Sectors for our financial market study. Each of these stocks is considered investment-worthy\footnoterule\footnote{Companies rated Baa or higher by Moody's and Standard \& Poor's (S\&P) are considered to be of high quality and deemed investment-worthy.} by both Moody's and S\&P. Recognizing the intrinsic challenge in accurately predicting stock prices as highlighted by Nguyen et al.\cite{nguyen2015sentiment}, we use blue-chip stocks in our research to anticipate upcoming stock price movements and volatility trends, as indicated by Feng et al.\cite{feng2019temporal} and Xu et al.~\cite{xu-cohen-2018-stock}.

In the domain of stock market research, two primary methodologies prevail: technical analysis and fundamental analysis. Technical analysis utilizes past stock prices to predict future trends\cite{lu2021cnn}. However, its heavy dependence on historical data can sometimes overlook sudden market changes due to unexpected events. Assuming a uniformly rational market behavior, this methodology can inadvertently create an echo chamber. This effect can cause trading signals to amplify themselves, eventually becoming disconnected from the actual economic context. Conversely, fundamental analysis integrates both price features and external information, including data from social media~\cite{cookson2020don} and search engines~\cite{wang2021stock}. Mao et al.\cite{10.1145/2392622.2392634} demonstrated an enhanced accuracy in forecasting the S\&P 500 closing price when integrating Twitter\footnote{Despite the recent rebranding of Twitter to ``X'', this article retains the use of its original name, ``Twitter''.} data into their model. While these data sources frequently reflect not only the financial market but also vital economic indicators, the prevailing research in fundamental analysis tends to emphasize the financial market, neglecting the symbiotic relationship between the broader economy and the stock market. Moreover, while existing models mainly center on forecasting trend shifts\cite{zhang2022transformer}, they often neglect the importance of the scale of these changes. In the realm of stock behavior, the magnitude of these shifts holds significant weight.

In this paper, we propose  ALERTA-Net:  \textbf{A}ttentional Tempora\textbf{L} Distanc\textbf{E} Awa\textbf{R}e Recurren\textbf{T} Neur\textbf{A}l Networks. To our best knowledge, it is the first paper to use the combination of social media, macroeconomic data and search engine information to predict both stock price movement and volatility. 
Our contributions can be summarized as follows:

\textbf{\textbullet\ Proposing a framework enabling the fusion of social media, macroeconomic factors and search engine data for stock movements
and volatility.} By integrating above information with our method, we can not only predict stock price movements but also efficiently extract information from stock market volatility. This allows us to provide advance warning of any unusual fluctuations in the stock market in the future.

\textbf{\textbullet\ Formulating a temporal distance-aware, multi-attention mechanisms
on multi-view market data.} The proposed ALERTA-Net shines in recognizing the dynamic, temporal distance-based relationships inherent within difference hidden states. Capitalizing the same day stock price movements, the model greatly amplifies its precision in predicting stock market volatility.

\textbf{\textbullet\ Validating the
effectiveness and efficiency of the proposed model via experiments
and comparisons}. We conduct experiments on one real-world dataset\footnote{Our dataset is available at https://github.com/hao1zhao/ALERTA-Net}. Both conventional methods and deep learning based methods for
stock market movements and volatility are selected for comparisons. Evaluations of various metrics are presented, illustrating the effectiveness of our proposed model.

%% file: sections/related_work.tex
\section{Related Work} 
Predictive methods for stock movement can be broadly categorized into two main types: technical analysis and fundamental analysis. While technical analysis relies exclusively on past price data to anticipate future trends, fundamental analysis adopts a more comprehensive approach, considering not only historical prices but also information from textual sources, economic indicators, financial metrics, and a myriad of both qualitative and quantitative aspects.

\textbf{Social Media}:
Numerous studies have explored stock market predictions using social media data. Contemporary research combines sentiment analysis with historical price data, extracting insights from platforms such as Yahoo's message board \cite{das2001yahoo}, blogs\cite{de2008can}, Twitter\cite{xu-cohen-2018-stock}, and Reddit\cite{long2023just}. This integration has revealed correlations with stock market trends.

\textbf{News and Search Engine}:
Exploring the influence of public news and user browsing habits, research has probed into how traders respond to news events. For instance, Xiong et al.\cite{xiong2015deep} regard Google trends and market data as catalysts for daily S\&P 500 variations. Other research, works like Bordino et al.\cite{bordino2012web} draw connections between search query volume and stock activity. More sophisticated techniques, like the hierarchical attention mechanisms introduced by Hu et al.\cite{hu2018listening}, extract news sequences directly from textual content to predict stock trends.

\textbf{Macroeconomic Indicators}:
Numerous studies pinpoint various economic elements influencing stock returns. Notably, Ferson et al. \cite{ferson1991variation} highlight the central role of interest rates in dictating stock returns. In addition, indicators like the relative T-Bill rate and the consumption-wealth ratio are underscored as crucial predictors by Jank et al. \cite{jank2012mutual}. Beyond these, economic markers like unemployment rates, inflation, and commodity prices also exert significant influence on stock returns, as affirmed by sources like \cite{hamilton2009causes} and \cite{kehoe2019asset}.

%% file: sections/prob_setup.tex
\section{Problem Setup}

 Let  $p = \{p^{1}, \ldots, p^{t}\}$ denote the stock's daily adjusted closing price. We  formulate the actual labels movement $y_m$ and volatility $y_v$ as follows:
\begin{equation}
y_m^t = \mathbb{1}(p^{t} - p^{t-1}),
\end{equation}
\begin{equation}
y_v^t = \mathbb{1}(p^{t} - p^{t-1})/p^{t-1}.
\end{equation}

The actual labels can be represented as $y_m = \{y_m^{1}, \ldots,y_m^{t}\}\in\mathbb{R}^{T}$ and $y_v\ = \{y_v^{1}, \ldots,y_v^{t}\}\in\mathbb{R}^{T}$.
Let ${X}^{T} = \{x^1, \ldots,x^t\} \in \mathbb{R}^{D\times T}$  represents the sequential input features (e.g., sentiment scores, stock adjusted prices) from the previous $T$ time-steps, where $D$ signifies the dimension of the features.
Since our goal is to utilize the sequence of features ${X^T}$ to predict the next time-step movements $\hat{y}_m$ and volatility $\hat{y}_v$ of blue-chip stocks. 
We can define our prediction functions $\hat{y}_m^{t} = f({X}^{T};{\Theta_1}), \hat{y}_v^{t} = g({X}^{T}$, $\hat{y}_m^{t};{\Theta_2})$, where $f$ the function  with parameters ${\Theta_1}$ aims to predict the movement of stock s at the next time-step from the sequential features ${X}^{T}$ and $g$ aims to predict the unusual fluctuation from both ${X}^{T}$ and $\hat{y}_m^{t}$.

In practical settings, by varying the time lag in extensive historical stock data, we can often produce numerous training examples. However, for clarity in presenting our proposed technique, we zero in on a distinct time lag for predicting both movement and volatility.
Additionally, our predictive models have assimilated the concept of adjusted stock prices\cite{xie2013semantic}. This ensures that our models discern authentic stock value changes driven by market factors, rather than getting influenced by artificial shifts arising from corporate decisions.

%% file: sections/model.tex
\section{Framework Components}

In this paper, we introduce a new framework, ALERTA-Net, which incorporates temporal distance-aware, multi-attention mechanisms for processing multi-view stock data. The overall architecture is illustrated in Figure 1. The data input \& preprocessing layer transforms both temporal and textual information into dense vectors. Then, our temporal distance-aware layer have the recurrent representation identifies hidden dependencies within the current stock data, based on past information. After that, distance-matrix context integrates these historical dependencies within the sequence of features ${X}^{T}$. Lastly, the predictions layer generates time-aware forecasts for the stock movements and volatility in the next time-step, thereby providing a complete and cohesive system for stock prediction.

\textbf{Data input \& preprocessing}. 
We designate the textual data extracted from Twitter as $\alpha$. To quantify the embedded sentiment within the Twitter text, we utilize the roBERTa-base sentiment model in combination with TweetNLP~\cite{camacho-collados-etal-2022-tweetnlp}, facilitating the generation of tweet sentiment scores denoted as $\hat{\alpha}$. Then, we concatenate $\hat{\alpha}$ with other relevant historical data,
yielding 
$E^T = \{e^1, \ldots,e^t\} \in \mathbb{R}^{D\times T}$, 
in which the feature dimension is 17.
Considering the fact that normalization provides a uniform scale to all features, 
thereby precluding any specific feature from dominating, we apply log normalization to $E^T$. For the purpose of avoiding numerical instability, we add a small constant $\epsilon = 1e-8$, the formula is as follows:

\begin{equation}
{X}^{T} = \log{E^T+\epsilon}.
\end{equation}

\textbf{Temporal distance aware (TDA)}.
In this layer, we propose a temporal-distance aware layer to enhance the modulation of conditional dependencies by allowing the model to access and directly attend to previous hidden states.
First, given the proficiency in handling long-term dependencies, recurrent neural network is extensively employed for sequential data processing~\cite{xu2018stock}. The general idea of recurrent unit is to recurrently project the input sequence into a sequence of hidden representations. At each time-step, the recurrent unit learns the hidden representations \(h_{t}\) by jointly considering the input \(x_{t}\) and previous hidden representation \(h_{t-1}\) to capture sequential dependency. To capture the sequential dependencies and temporal patterns in the historical stock features, an GRU\cite{cho2014learning} recurrent unit is applied to map $ \{x^{1}, \ldots, x^{t}\}$  into hidden representations $ \{h^{1}, \ldots, h^{t}\}\in\mathbb{R}^{U\times T}$, with the dimension of $U$. 

Instead of just using the immediate previous hidden state $h^t$ to update the current hidden state $h^{t+1}$, we are now considering a weighted sum of all previous hidden states. This not only enable the model to place greater emphasis on the impact of recent events on the stock market but also to consider a longer history. We utilized a temporal distance to represent the interval between two time steps.
\begin{equation}
w^{i} = \frac{1}{t - i + 1},
\end{equation}
where the weight $w^i$ for a hidden state at time $i$ based on its temporal distance from the current time $t$. By adding 1 to the denominator, we ensure that even the most recent hidden state gets a weight, preventing division by zero.

Then, we apply weights to the hidden states of the GRU:
\begin{equation}
c^{t} = \text{GRU}\left(x^{t}, \sum_{i=1}^{t} w^i \cdot h^{i}\right),
\end{equation}
each hidden state $h^i$ is multiplied by its respective weight $w^i$, and the results are summed up. The result is a context state $c^t$ that captures the weighted influence of all previous states.

\textbf{Prediction Layer}.
Instead of directly predicting stock movement, denoted as \( {y}_{m}^{t} \), and volatility \( {y}_{v}^{t} \), we first concatenate the context \( c^t \) with the previous hidden state \( h^{t} \) in the fusion layer. We then use the cross-entropy function as the optimizer for stock price movement prediction. Following this, we concatenate the last hidden state \( h^{t} \), context \( c^{t} \), and the output \( {y}_{m}^{t} \) to predict volatility \( {y}_{v}^{t} \). For this prediction, we utilize the binary cross-entropy with logits as the loss function.

%% file: sections/experiment.tex
\section{Experiment}

\subsection{Dataset Description}

Our dataset coalescest three main components: Historical price data, Twitter data and macroeconomics data. 

\textbf{Yahoo Finance}. We sourced historical data from Yahoo Finance, which monitored the trajectory of 41 blue-chip stocks between June 1, 2020, and June 1, 2023. To hone our prediction objectives, we designated a threshold range spanning from -0.5\% to 0.5\% to filter out negligible shifts. While Baker et al.\cite{baker2021triggers} suggest daily stock price alterations exceeding 2.5\% are deemed notable, our model's aim is to forecast atypical volatility. Aligning with Ding et al.'s insights \cite{ding2007private} on distinct stock fluctuation parameters, we've established a loftier standard, recognizing a 5\% swing as an outlier. As a result, we categorize samples with variances below 5\% as 0 and those at or above 5\% as 1.

\textbf{Twitter}. During the same date range, we included approximately 7.8 million tweets, gathered via Twitter's official API at a sampling rate of 10\%. We were particular in our selection of tweets: they needed to contain at least one cashtag and had to be posted within standard U.S. trading hours, from 9 am to 4:30 pm. We recognized the significant influence that Twitter volume has on stock trading, a fact underscored by Cazzoli et al. \cite{cazzoli2016large}. Thus, we ensured that our model's input parameters included the daily count of processed Twitter posts.

\textbf{Google Trends \& Federal Reserve Economic Data}. We engaged in targeted searches on Google and Federal Reserve Economic Data, using carefully curated keywords originating from the ``Outline of Economics'' Wikipedia page. In light of the disparate update intervals across various indicators, we broke down each data pull into smaller segments, ensuring normalization of the data across these windows.

\subsection{Baseline Methods and Evaluation Metrics}
We evaluate our model's effectiveness by comparing it with DP-LSTM\cite{li2019dp}, a renowned stock movement prediction network by using financial data. 

Other benchmarks employed in our study include Extreme Gradient Boosting\cite{chen2016xgboost}, Attention-based LSTM\cite{wang2016attention}, and GRU\cite{cho2014learning}. Following similar procedures in (Xu et. al~\cite{xu2018stock}; Zhang et. al~\cite{zhang2022transformer}), we report our results in terms of Accuracy (Acc.) and Matthews Correlation Coefficient (MCC). Given that data points involving stock price changes greater than 5\% only constitute a minor portion of our dataset, we've also chosen to utilize the Area Under the ROC Curve (AUC) as our performance metric in order to achieve a more robust and realistic evaluation.

\begin{table}[ht]
\centering
\parbox{\columnwidth}{
\normalsize
    TABLE I : Classification performance of ALERTA-NET and baseline methods, measured with the accuracy (Acc.), the Matthews Correlation Coefficient (MCC) and Receiver Operating Characteristic (ROC). The best is in bold, our ALERTA-NET shows the best performance in all  evaluation metrics.
    \par\vspace{5pt} 
}
{\footnotesize 
\begin{tabularx}{\columnwidth}{@{}l*{5}{>{\centering\arraybackslash}X}@{}}
\toprule
\multicolumn{1}{c}{\multirow{2}{*}{Models}} &
  \multicolumn{2}{c}{Movement} &
  \multicolumn{3}{c}{Volatility} \\ 
\cmidrule(r){2-3} \cmidrule(r){4-6} 
\multicolumn{1}{c}{} &
  Acc. & 
  MCC &
  Acc. &
  MCC &
  AUC \\ 
\midrule

XGBoost & 0.5030 & 0.01461 & 0.5240 & -0.02275 & 0.5561 \\
GRU  & 0.5121 & 0.04217 & 0.5633 & 0.06241 & 0.5993 \\
DP-LSTM & 0.5036 & 0.00741 & 0.5684 & 0.05297 & 0.5890 \\
AT-LSTM & 0.5165 & 0.03779 & 0.5886 & 0.07028 & 0.5850 \\
ALERTA-Net      & \textbf{0.5238} & \textbf{0.05179} & \textbf{0.6136} & \textbf{0.07465} & \textbf{0.6197}\\ 
\bottomrule
\end{tabularx}}
\end{table}

\begin{table}[ht]
\centering
\parbox{\columnwidth}{
\normalsize
    TABLE II : An ablation study of ALERTA-Net, where ALERTA-Net(P) is baseline without any tweet and macroeconomic data. ALERTA-Net(S) is only relies on sentiment data. ALERTA-Net(W/O M) is without macroeconomic data. Both tweet sentiment and macroeconomic data improve the performance of the framework.
    \par\vspace{4pt} 
}
\label{tab2}

{\footnotesize 
\begin{tabularx}{\columnwidth}{@{}l*{5}{>{\centering\arraybackslash}X}@{}}
\toprule
\multicolumn{1}{c}{\multirow{2}{*}{Models}} &
  \multicolumn{2}{c}{Movement} &
  \multicolumn{3}{c}{Volatility} \\ 
\cmidrule(r){2-3} \cmidrule(r){4-6} 
\multicolumn{1}{c}{} &
  Acc. &
  MCC &
  Acc. &
  MCC &
  AUC \\ 
\midrule
ALERTA-Net(P)      & 0.4988 & 0.0016 & 0.5238 & 0.07168 & 0.6019 \\
ALERTA-Net(S)          & 0.4952 & 0.0092 & 0.5150 & 0.03213 & 0.5753 \\
ALERTA-Net(W/O M)  & 0.5114 & 0.0378 & 0.5247 & 0.03992 & 0.5989 \\
\midrule
ALERTA-Net                  & \textbf{0.5238} & \textbf{0.05179} & \textbf{0.6136} & \textbf{0.07465} & \textbf{0.6197}   \\ 
\bottomrule
\end{tabularx}
} 
\end{table}

\subsection{Results}
The performances of our proposed models and the established baselines are detailed in TABLE \MakeUppercase{\romannumeral 1}. AT-LSTM is observed to be the superior baseline model in  terms of accuracy and MCC  for movement prediction, while GRU shows an outstanding performance in the AUC (Area Under the Curve) and MCC for volatility prediction. ALERTA-Net surpasses both of these models by significant margins. In terms of accuracy, ALERTA-Net achieves a score of 0.5238 and 0.6136, outperforming GRU and AT-LSTM by 1.4\% and 4.2\%  respectively. Additionally, for MCC, ALERTA-Net outperforms GRU and AT-LSTM by 22.8\% and 6.2\% respectively, and outshines both in AUC by a margin of 3.4\%. Overall, these results reinforce the effectiveness of our proposed model ALERTA-Net.

\subsection{Ablation Study}
To conduct an in-depth analysis of the core components of ALERTA-Net, we have constructed three variations alongside the fully-loaded model. Each variant is specifically tailored to handle certain types of input data: ALERTA-Net(P) solely relies on closed price data, ALERTA-Net(S) exclusively processes Twitter-derived sentiment data, while ALERTA-Net(W/O M) incorporates both price and sentiment data, but omits macroeconomic information. As shown in TABLE \MakeUppercase{\romannumeral 2}, our ablation study revealed that incorporating macroeconomic data significantly enhances the predictive capabilities of the model for stock movement and volatility to varying degrees.

%% file: sections/conclusion.tex
\section{Conclusion}
We introduced ALERTA-Net, a deep generative neural network architecture, to showcase the efficacy of combining  search engine data, macro-economy data, and social media data when trying to predict stock movements and volatility. We tested our model on a new comprehensive dataset and showed it performs better than strong baselines. In future studies, we plan to enhance accuracy by integrating a variety of text and audio sources, including earnings calls and financial reports.